# MyCaffe: A Complete C# Re-Write of Caffe with Reinforcement Learning

David W. Brown
daveb@signalpop.com
SignalPop LLC.

10/4/2017

(Updated 9/23/2018)

**Abstract**

*Over the past few years Caffe* [1], *from Berkeley AI Research, has gained a strong following in the deep learning community with over 15k forks on the github.com/BLVC/Caffe site. With its well organized, very modular C++ design it is easy to work with and very fast. However, in the world of Windows development, C# has helped accelerate development with many of the enhancements that it offers over C++, such as garbage collection, a very rich .NET programming framework and easy database access via Entity Frameworks* [2]. *So how can a C# developer use the advances of C# to take full advantage of the benefits offered by the Berkeley Caffe deep learning system? The answer is the fully open source, 'MyCaffe' for Windows .NET programmers. MyCaffe is an open source, complete re-write of Berkeley's Caffe, in the C# language.*

*This article describes the general architecture of MyCaffe including the newly added MyCaffeTrainerRL for Reinforcement Learning. In addition, this article discusses how MyCaffe closely follows the C++ Caffe, while talking efficiently to the low level NVIDIA CUDA hardware to offer a high performance, highly programmable deep learning system for Windows .NET programmers.*

## Introduction

The goal of MyCaffe is not to replace the C++ Caffe, but instead to augment the overall platform by expanding its footprint to Windows C# programmers in their native programming language so that this large group of software developers can easily develop Caffe models that both the C++ Caffe and C# MyCaffe communities benefit from. MyCaffe also allows Windows developers to expand MyCaffe with their own new layers using the C# language.

### C++ Caffe and C# MyCaffe Comparison Table

|  | **C++ Caffe** | **C# MyCaffe** |
|---|---|---|
| All vision layers | x | x |
| All recurrent layers | x | x |
| All neuron layers | x | x |
| All cuDnn layers | x | x |
| All common layers | x | x |
| All loss layers | x | x |
| All data layers | x | All but 2 |
| All auto tests | x | x |
| All solvers | x | x |
| All fillers | x | x |
| Net object | x | x |
| Blob object | x | x |
| SyncMem object | x | x |
| DataTransform object | x | x |
| Parallel object(s) | x | x |
| NCCL (Nickel) | x | x |
| Caffe object | x | CaffeControl |
| Database | Levledb, Imdb | MSSQL (+Express) |
| Low-level Cuda | C++ | C++ |
| ProtoTxt | Text | Compatible |
| Weights | Binary | Compatible |
| GPU Support | 1-8 gpus | 1-8 gpus* |
| CPU Support | x | GPU only |

*Table 1 Support Comparison*

\* Multi-GPU support only on headless GPUs.

From Table 1 (above) you can see that MyCaffe is a very close match to the C++ Caffe, with the main differences being in the database support and the way that MyCaffe accesses the low-level C++ CUDA code. In addition, MyCaffe is designed to only run on GPU bases systems and supports a wide range of NVIDIA cards from the NVIDIA GeForce 750ti on up to multi-GPU Tesla based systems running in TCC mode.

Like the C++ Caffe, MyCaffe can be complied to run using either a 'float' or 'double' as its base type.

## MyCaffe General Architecture

Four main modules make up the MyCaffe software: MyCaffe Control, MyCaffe Image Database, Cuda Control and the low-level Cuda C++ Code DLL.

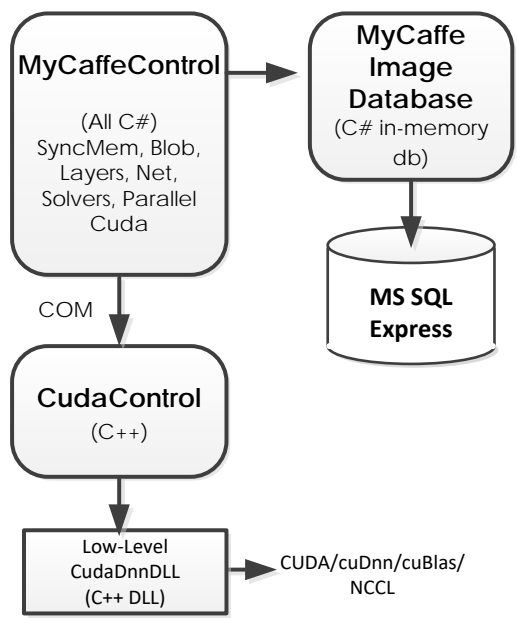

*Table 2 MyCaffe Architecture*

The MyCaffe Control is the main interface to software that uses MyCaffe. It mainly plays the role of the 'caffe' object in C++ Caffe but does so in a way more convenient for C# Windows programmers for the control is easily dropped into a windows program and used.

The MyCaffe Image Database is an in-memory database used to load datasets, or portions of datasets, from a Microsoft SQL (or Microsoft SQL Express) database into memory. In addition to providing the typical database functionality, the MyCaffe Image Database adds several useful features to deep learning, namely: label balancing [3] and image boosting [4]. When using label balancing the up-sampling occurs at the database. The database itself organizes all data by label when loaded, thus allowing a random selection of the label group first, and then a random selection of the image from the label group, so as to ensure that labels are equally represented when training and not skew training toward one label or another. Image boosting allows the user to mark specific images with a higher boost value to increase the probability that the marked image is selected during training. These are optional features that we have found helpful when training on sparse datasets that tend to have imbalanced label sets.

The Cuda Control is a C++ COM Control that supports COM/OLE Automation communication. Using the COM support built into C#, the MyCaffe Control communicates to the Cuda Control seamlessly by passing parameters and a function index, which is then used internally to call the appropriate CUDA based function within the Low-Level CudaDnn DLL. All parameters are converted from the COM/OLE Automation format and into native C types within the Cuda Control thus allowing the Low-Level CUDA C++ Code to focus on the low-level programming such as calling CUDA kernels and/or calling cuDnn and NCCL functions.

The Low-Level CudaDnn DLL (different from, but uses the cuDnn [5] library from NVIDIA) contains all of the low-level functionality such as the low level MyCaffe math functions, calls to cuDnn, calls to NCCL and several low level t-SNE [6] and PCA [7] functions provided for speed. In order to manage the CUDA memory, CUDA Streams and special object pointers, such as the Tensor Descriptors used by cuDnn, the CudaDnn DLL uses a set of look-up tables and keeps track of each and every memory allocation and special pointer allocated. A look-up index into each table acts as a handle that is then passed back on up to, and used by, the C# code within the MyCaffe Control. Thanks to the way GPU's operate with a (mainly) separate memory space from the CPU, a look-up table works well for once the memory is loaded on the GPU, you generally want to keep it there for as long as possible so as to avoid the timing hit caused by transferring between the GPU and host memory.

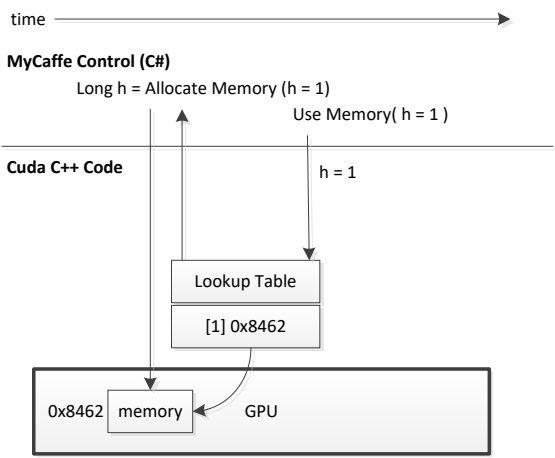

*Table 3 Fast Lookup Table*

MyCaffe also employs this same concept to manage all special data pointers used by the Cuda and cuDnn libraries, where each pointer to an allocated descriptor, or stream, is also placed in its own look-up table that is referenced using an index (or handle) up in the MyCaffe Control software. To support some sub-systems, such as NCCL, t-SNE and PCA, internal objects are allocated to manage the sub-system and stored in a look-up table thus allowing the CudaDnn DLL to manage the state of these subsystems while they are being used.

## Benchmarks

In deep learning, speed is important. For that reason, we have run the MyCaffe on several Windows configurations to show how well the technology stacks up[1].

The following benchmarks were run on an i7-6950 10-core 3.00GHZ Dell/Alienware Area-51 R2 Computer running Windows 10 Pro with 64GB of memory loaded with an NVIDIA Titan Xp (Pascal) GPU running in TCC[2] mode using CUDA 9.2.148/cuDNN 7.2.1 (9/12/2018).

| Model | GPU Memory Used | Image Size | Batch Size | Fwd/Bwd Average Time (ms.) |
|---|---|---|---|---|
| GoogleNet | n/a | 227x227 | *128[3] est* | *610 ms* |
| GoogleNet | 7.76 GB | 227x227 | 64 | 305 ms |
| GoogleNet | 4.55 GB | 227x227 | 32 | 160 ms |
| GoogleNet | 2.92 GB | 227x227 | 16 | 75 ms |
| VGG-16 | 7.91 GB | 227x227 | 32 | 350 ms |
| VGG-16 | 5.71 GB | 227x227 | 16 | 196 ms |
| GoogleNet | 3.29 GB | 56x56 | 24 | 67 ms |
| AlexNet | 1.02 GB | 32x32 | 128 | 45 ms |

Comparison benchmarks (run in 2014) show that C++ Caffe using an older version of cuDnn running GoogleNet on an NVIDIA K40c with a 128 batch size had a running total time of 1688.8 ms. for the forward + backward pass [8].

But how does MyCaffe stack up on lower-end GPU's such as the 1060 typically found in laptops or the $170 NVIDIA 1050ti? The following benchmarks show what these more standard PC systems can do, as well.

The following benchmarks were run on an i7 6700HQ 6-core 2.60GHZ Alienware 15 Laptop running Windows 10 with 8 GB of memory loaded with an NVIDIA GTX 1060 GPU in WDM[4] mode with CUDA 9.2.148/cuDNN 7.2.1.

| Model | GPU Memory Used | Image Size | Batch Size | Fwd/Bwd Average Time (ms.) |
|---|---|---|---|---|
| GoogleNet | 4.20 GB | 56x56 | 24 | 180 ms |
| AlexNet | 1.61 GB | 32x32 | 128 | 77 ms |

---

[1] The SignalPop AI Designer version 0.9.2.188 using the LOAD_FROM_SERVICE mode was used for all tests.
[2] Run on GPU with monitor plugged in.
[3] Estimated by multiplying batch 64 timing x2.
[4] Run on GPU with monitor connected.

The following benchmarks were run on an Intel-Celeron 2.90GHZ system running Windows 10 Pro with 12 GB of memory loaded with an NVIDIA 1050 GPU running in WDM[5] mode with CUDA 9.2.148/cuDNN 7.2.1.

| Model | GPU Memory Used | Image Size | Batch Size | Fwd/Bwd Average Time (ms.) |
|---|---|---|---|---|
| GoogleNet | 3.84 GB | 56x56 | 24 | 285 ms |
| AlexNet | 1.46 GB | 32x32 | 128 | 125 ms |

As expected, the NVIDIA 1060 and 1050 systems run slower than the Tesla TCC based systems but are still able to run very complicated models such as GoogleNet - just with a smaller batch size.

### Compatibility

Since our goal is to expand the target market for the C++ Caffe platform to the general C# Windows programmer, we have strived to make sure that MyCaffe maintains compatibility with the C++ Caffe platform prototxt scripts and binary weight file formats. To do this, MyCaffe performs its own parsing of the prototxt files into internal C# parameter objects that closely match those generated by Google's ProtoTxt software. In addition each parameter object is designed to print itself out in the very same ProtoTxt format from which it was parsed. Using this methodology allows MyCaffe to maintain compatibility with the same prototxt files used by the C++ Caffe.

With regard to the weigh files, MyCaffe stores weight files using the same Google ProtoBuf binary format as the C++ Caffe allowing MyCaffe to use existing *.caffemodel* files created using the C++ Caffe.

### Reinforcement Learning Support

MyCaffe recently added support for policy gradient reinforcement learning [9] which runs on a Cart-Pole simulation written in C# inspired by the version by OpenAI [10] and originally created by Sutton et al. [11] [12]. A new component called the MyCaffeTrainerRL provides the basic architecture for training a MyCaffe component using the reinforcement learning style of training.

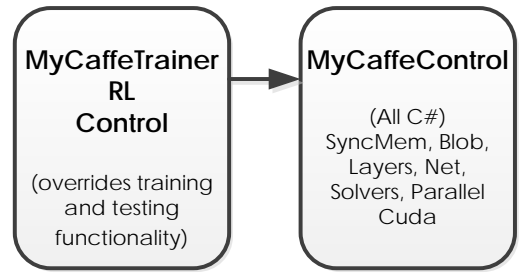

*Figure 1 MyCaffeTrainerRL uses MyCaffe*

When using the MyCaffeTrainerRL, the same MyCaffe solver, network and model are used, but trained with a reinforcement learning method of training. The main requirements are that the model use the MemoryData layer for data input, and the MemoryLoss layer to calculate the loss and gradients. For example the Sigmoid based model below uses both where the Sigmoid layer produces a probability used to determine one of two actions to take.

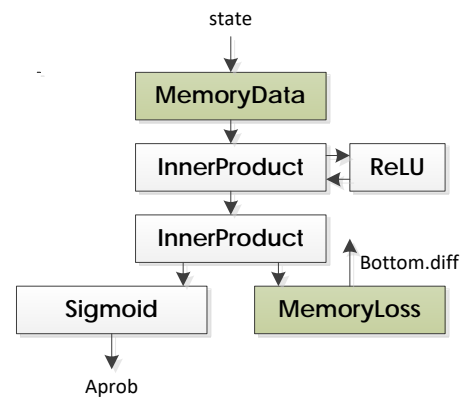

*Figure 2 Policy Gradient Reinforcement Learning Model with Sigmoid*

Alternatively, a Softmax based model may be used as well when more than two actions are needed.

---

[5] Run on GPU with monitor connected.

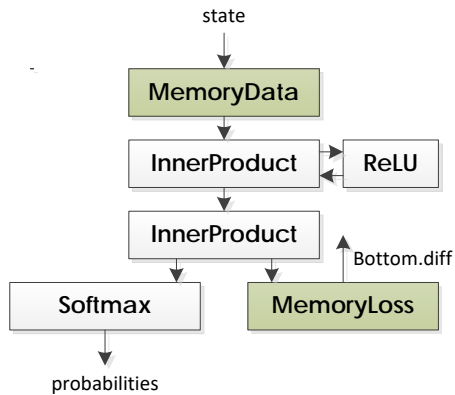

*Figure 3 Policy Gradient Reinforcement Learning Model with Softmax*

When training either model, the type of model is detected by the MyCaffeTrainerRL which hooks into the loss functionality of the MemoryLoss layer and takes care of calculating the losses and setting the gradients on the *Bottom.diff* of the layer attached to the Memory Loss layer. The back-propagation then propagates the *Bottom.diff* on up the network, from bottom to top, as it does with any other model.

For more details on how this works, see Appendix A (Sigmoid) or Appendix B (Softmax).

## Summary

It is great to see fantastic software such as Caffe out in the open source community. As a thank you, we wanted to contribute back to that same community with the open-source 'MyCaffe' project to help an even larger group of programmers use this great technology. And now with the MyCaffeTrainerRL, it is even easier to create reinforcement learning solutions with MyCaffe.

For more information on the C++ Caffe, see the Berkeley AI Research site at http://caffe.berkeleyvision.org/.

For more information on the Windows version of MyCaffe written in C#, see us on GitHub at https://github.com/mycaffe. For easy integration into your existing C# applications, just search for MyCaffe on NuGet or go to https://www.nuget.org/packages?q=MyCaffe.

And for more information on innovative products that make visually designing, editing, training, testing and debugging your AI models easier, see us at https://www.signalpop.com.

# Appendix A – Sigmoid based Reinforcement Learning with MyCaffe

The new MyCaffeTrainerRL extends MyCaffe by adding the ability to easily train policy gradients with reinforcement learning. When using the MyCaffeTrainerRL with two actions, the following model architecture is used.

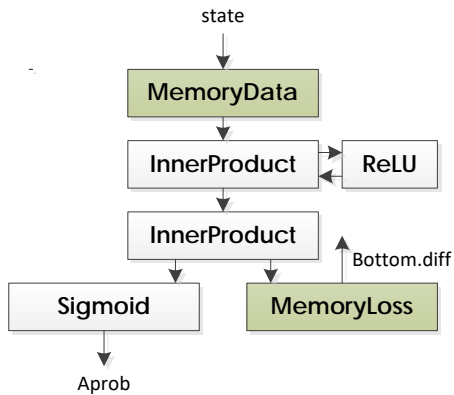

*Figure 4 Sigmoid based Policy Gradient Reinforcement Loss Model*

Each model used with the MyCaffeTrainerRL, is required to use both the MemoryData layer for input, and the MemoryLoss layer to calculate the gradients – the MyCaffeTrainerRL actually takes care of adding the input to the MemoryData layer and automatically calculates the loss and gradients by hooking into the MemoryLoss layer.

The gradients are then fed back up through the *Bottom.diff* of the layer feeding into the MemoryLoss layer, which in this case is the bottom InnerProduct layer.

The main limitation of a Sigmoid based model is that it only supports two actions. For more than two actions, you will want to use a Softmax based model. The Sigmoid model is faster than the Softmax model, so if you only have two actions, it is recommended – see Appendix B (below) to see how the Softmax based model works.

## Detailed Walk Through

The diagram below shows how each step of the MyCaffeTrainerRL works to provide the reinforcement learning with a Sigmoid based model. During this process, the following steps take place:

0.) At the start, the Environment (such as Cart-Pole) is reset, which returns our initial state – *state0*.
1.) The *state0* is fed through the network (and its Model) in a forward pass…
2.) …to produce the probability used to determine the action - the probability produced is called *Aprob*.
3.) The action is set as follows: *action = (random < Aprob) ? 0 : 1*, which sets the action to zero if the random number is less than *Aprob* and to 1 otherwise.
4.) The *action* is then fed to the Environment, which is directed to run the action in the next Step of the simulation.
5.) The new Step produces a new state, *state1* and a *reward* for taking the *action* (in the case of Cart-Pole, after taking the *action*, the reward is set to 0 if the cart runs off the track or the pole angle exceeds 20 degrees, and to 1 if the pole is still balancing). The previous state (*state0*), the *action* taken, the *Aprob* used to calculate the *action* and the *reward* for running the action are all stored in a Memory that collectively contains a full 'experience' once the simulation completes the round.
6.) If the simulation is **not** complete, the *state0* is set to *state1* and we continue back up to step 1 above.
7.) Once the simulation is complete (e.g. the Environment at step 4 returns *done* = true) the episode is processed by training the network on it.
8.) At the start of training, the initial gradient is calculated as follows: *Dlogps = (action==0) ? 1-Aprob : 0 – Aprob*, which will help move the weights toward what the action should be [13] [14] (for more on this see Sigmoid Gradient Calculation below).

9.) Next, we calculate the *discounted rewards*, which are discounted backwards in time so as to give older steps higher weighting than newer steps, which helps encourage larger and larger episodes.
10.) The *discounted rewards* are multiplied by the policy gradients (*Dlogps*) to produce a set of modulated gradients.
11.) The modulated gradients are set as the *Bottom.diff* in the layer connected to the MemoryLoss layer (which in this case is the bottom InnerProduct layer), and then the backward pass back propagates the diff on up through the network. NOTE: the gradients of the network are accumulated until we hit a *batch_size* number of episodes, at which time the gradients are applied to the weights.
12.) The Environment is reset to start a new simulation and we continue back up to step 1 to repeat the process.

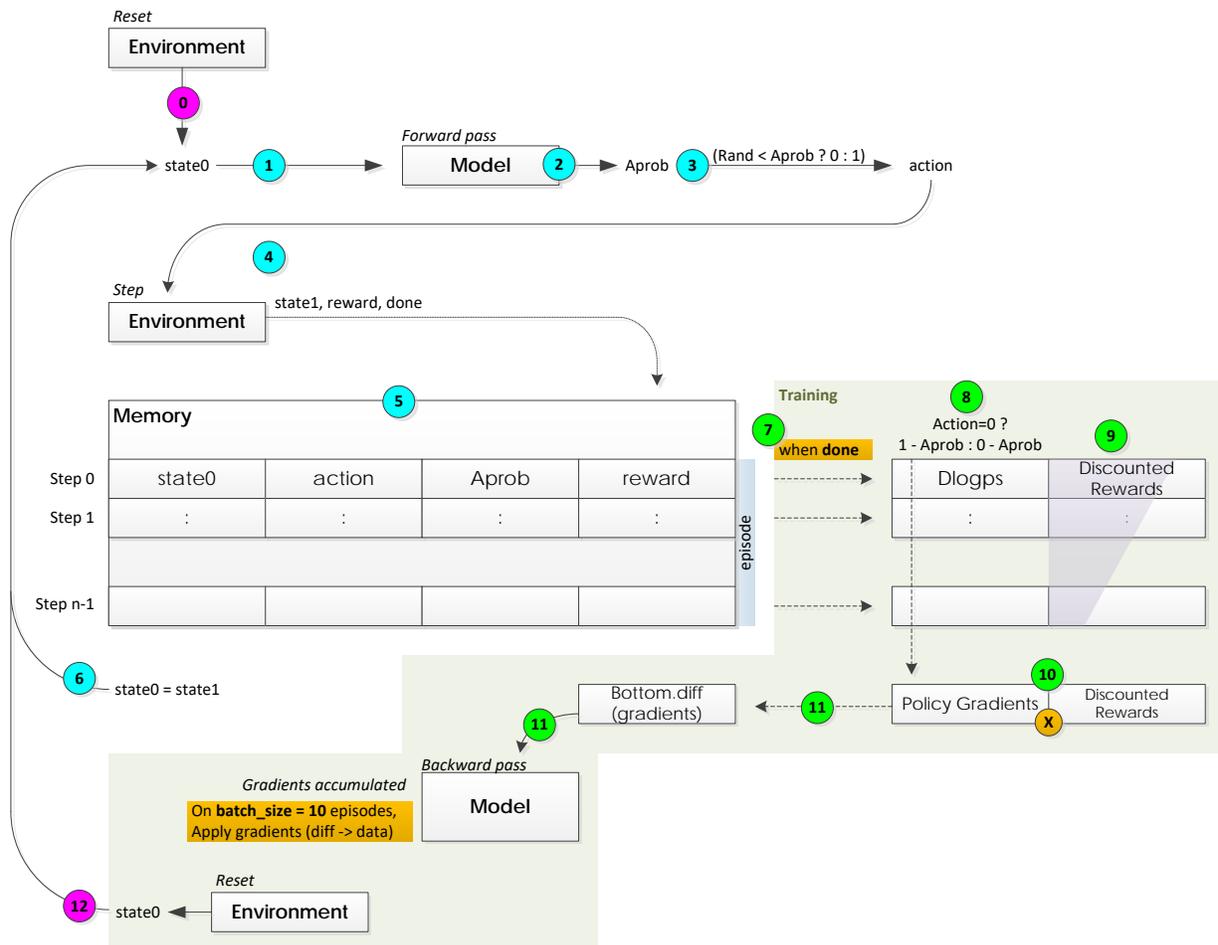

*Figure 5 Reinforcement Learning Process with Sigmoid*

## Sigmoid Gradient Calculation

The key to the Sigmoid based policy gradient reinforcement learning is in the calculation of the gradients applied to the *Bottom.diff*. There are three formulas that perform this task:

*action = (random < Aprob) ? 0 : 1*

*Dlogps = (action==0) ? 1-Aprob : 0 – Aprob*        // determine the gradient

*Bottom.diff = Dlogps * discounted rewards*        // modulate the gradient

How does this actually work? The following table shows what happens to the data in four examples (rows) as the data moves through the network.

| | col. 1 | col. 2 | col. 3 | col. 4 | col. 5 | col. 6 | col. 7 | col. 8 | col. 9 | |
|---|---|---|---|---|---|---|---|---|---|---|
| | Aprob (ap) | (random<ap)?0:1 Action (a) | (a=0)?1-ap:0-ap Dlogps | diff | old weight | new weight | data | old result | new result | |
| row 1 | 0 | 1 | 0 | 0 | 1 | 1 | 0 | 0 | 0 | |
| row 2 | 0.1 | 1 | -0.1 | 0.1 | 1 | 0.9 | 0.1 | 0.1 | 0.09 | decreased toward 0 |
| row 3 | 0.9 | 0 | 0.1 | -0.1 | 1 | 1.1 | 0.9 | 0.9 | 0.99 | increased toward 1 |
| row 4 | 1 | 0 | 0 | 0 | 1 | 1 | 1 | 1 | 1 | |

result = weight * data

*Figure 6 Sigmoid Gradient Calculation*

Staring with column 1, let's trace through what actually happens to the data in the network, after which, you will hopefully better understand how the three functions above actually work to move the weights toward values that calculate the results we want.

At **column 1**, we first calculate *Aprob* by running the state through a network forward pass which is shown in steps 1 & 2 above.

Next, at **column 2**, the *action* is calculated using *action = (random < Aprob) ? 0 : 1*, also shown in step 3 above.

At **column 3**, the initial gradient *Dlogps* is calculated with *Dlogps = (action==0) ? 1-Aprob : 0 – Aprob*, also shown in step 8 above.

At **column 4**, we multiply -1 by *Dlogps* to compensate for the fact that MyCaffe (and Caffe) subtract the gradients (*Bottom.diff*) from the weights.

At **column 5** in our example, let's assume that the weight value is set to 1.0.

The new weight in **column 6** is calculated with *new weight = old weight – diff*, and this actually occurs during the Solver ApplyUpdate shown in step 11 above.

To see what impact the weight update actually had, let's assume that our input data in **column 7** is the same *Aprob* that we previously had as output, for we are trying to drive our weights to values that eventually produce the ground truths shown in rows 1 and 4.

Comparing the new result in **column 9**, produced with *new result = data * new weight* with the old result in **column 8**, we can see that our new result values are indeed moving closer to our ground truths. In row 2, our old result of 0.1 moves to 0.09 – a little closer to the ground truth *Aprob* value of 0 (which would always produce an action of 1), and in Row 3, our old result of 0.9 moves to 0.99 – a little closer to the ground truth *Aprob* value of 1 (which would always produce an action of 0).

As shown above, the *Dlogps* does indeed move our weights in the direction that also moves the final result toward our ground truth values.

# Appendix B – Softmax based Reinforcement Learning with MyCaffe

The MyCaffeTrainerRL that extends MyCaffe, by adding the ability to easily train policy gradients with reinforcement learning, also supports Softmax based models. When using the MyCaffeTrainerRL with more than two actions, the following general model architecture is used.

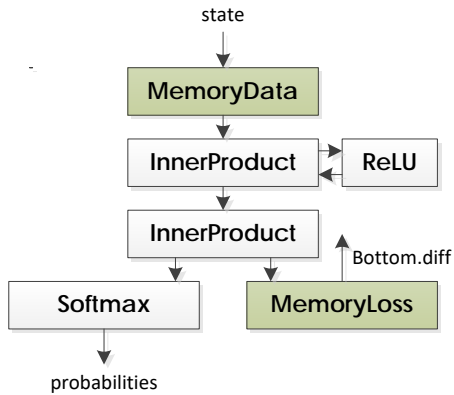

*Figure 7 Softmax based Policy Gradient Reinforcement Loss Model*

Just like the Sigmoid model, Softmax models are also required to use both the MemoryData layer for input, and the MemoryLoss layer to calculate the gradients – the MyCaffeTrainerRL actually takes care of adding the input to the MemoryData layer and automatically calculates the loss and gradients by hooking into the MemoryLoss layer.

With the Softmax model, the MyCaffeTrainerRL internally uses a SoftmaxCrossEntropyLoss layer to calculate the initial gradients that are then fed back up through the *Bottom.diff* of the layer feeding into the MemoryLoss layer, which in this case is the bottom InnerProduct layer.

Although slightly slower than the Sigmoid based model, Softmax based models easily support more than two actions required by some problems.

## Detailed Walk Through

The diagram below shows how each step of the MyCaffeTrainerRL works to provide the reinforcement learning with a Softmax based model. During this process, the following steps take place:

0.) At the start, the Environment (such as Cart-Pole) is reset, which returns our initial state – *state0*.
1.) The *state0* is fed through the network (and its Model) in a forward pass…
2.) …to produce a set of probabilities (one per action) used to determine the action – these our outputs of the Softmax layer.
3.) The probabilities returned by the Softmax layer are treated as a probability distribution used to determine the actual action.
4.) The *action* is then fed to the Environment, which is directed to run the action in the next Step of the simulation.
5.) The new Step produces a new state, *state1* and a *reward* for taking the *action* (in the case of Cart-Pole, after taking the *action*, the reward is set to 0 if the cart runs off the track or the pole angle exceeds 20 degrees, and to 1 if the pole is still balancing). The previous state (*state0*), the *action* taken and the *reward* for running the action are all stored in a Memory that collectively contains a full 'experience' once the simulation completes the round.
6.) If the simulation is **not** complete, the *state0* is set to *state1* and we continue back up to step 1 above.
7.) Once the simulation is complete (e.g. the Environment at step 4 returns *done* = true) the episode is processed by training the network on it.

8.) At the start of training, the initial gradient is calculated using an internal SoftmaxCrossEntropyLoss layer which produces the initial gradients that help move the weights toward what the action should be [13] [14] (for more on this see Softmax Gradient Calculation below).
9.) Next, we calculate the *discounted rewards*, which are discounted backwards in time so as to give older steps higher weighting than newer steps, which helps encourage larger and larger episodes.
10.) The *discounted rewards* are multiplied by the policy gradients (*SoftmaxCrossEntropyLoss gradients*) to produce a set of modulated gradients.
11.) The modulated gradients are set as the *Bottom.diff* in the layer connected to the MemoryLoss layer (which in this case is the bottom InnerProduct layer), and then the backward pass back propagates the diff on up through the network. NOTE: the gradients of the network are accumulated until we hit a *batch_size* number of episodes, at which time the gradients are applied to the weights.
12.) The Environment is reset to start a new simulation and we continue back up to step 1 to repeat the process.

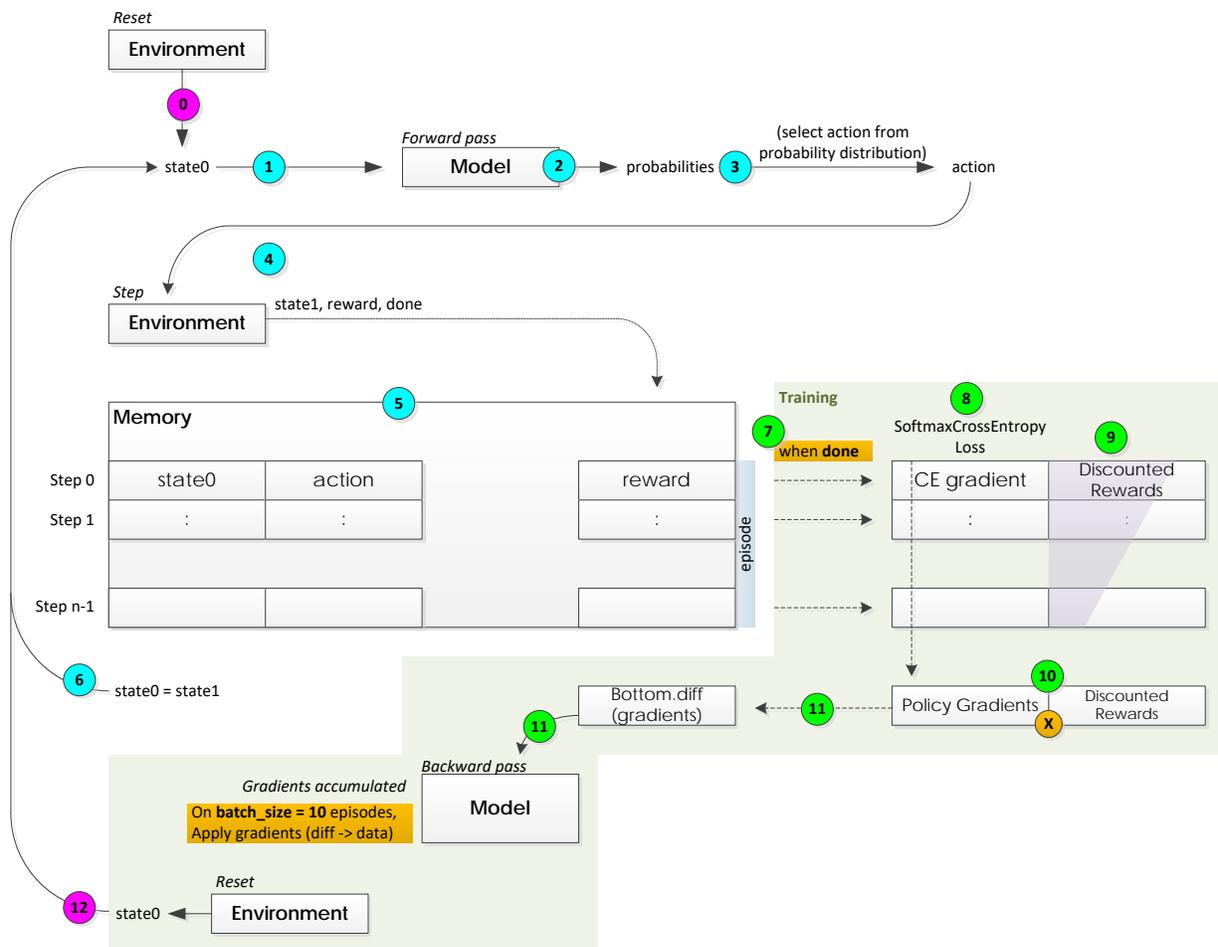

*Figure 8 Reinforcement Learning Process with Softmax*

Note, the training process of both the Softmax based and Sigmoid based models are basically the same except in how the gradients are calculated. The Softmax uses the SoftmaxCrossEntropyLoss to calculate the initial gradients, whereas the Sigmoid calculates the initial gradients directly (e.g. *Dlogps*).

## Softmax Gradient Calculation

The key to the Softmax based policy gradient reinforcement learning is in the calculation of the gradients applied to the *Bottom.diff*. There are three formulas that perform this task:

*softmax = calculated from action logits.*

*Dlogps = softmax - target*                        *// determine the gradient*

*Bottom.diff = Dlogps * discounted rewards*      *// modulate the gradient*

How does this actually work? The following table shows what happens to the data in four examples (rows) as the data moves through the network.

| | col. 1 | | col. 2 | | col. 3 | col. 3 | col. 4 | | col. 5 | col. 6 | | col. 7 | col. 8 | | col. 9 | col. 10 | col. 11 |
|---|---|---|---|---|---|---|---|---|---|---|---|---|---|---|---|---|---|
| | probabilities (from Softmax) | | targets | | softmax-target | softmax-target | | | old | new | | data | old | | | rew result | new result |
| | action 0 | action 1 | action 0 | action 1 | gradient 0 | gradient 1 | diff 0 | diff 1 | weight | weight | | | result | | | action 0 | action 1 |
| row 1 | 0 | 1 | 0 | 1 | 0 | 0 | 0 | 0 | 1 | 1 | 1 | 0 | 1 | 0 | 1 | 0 | 1 |
| row 2 | 0.1 | 0.9 | 0 | 1 | 0.1 | -0.1 | 0.1 | -0.1 | 1 | 0.9 | 1.1 | 0.1 | 0.9 | 0.1 | 0.9 | 0.09 | 0.99 |
| row 3 | 0.9 | 0.1 | 1 | 0 | -0.1 | 0.1 | -0.1 | 0.1 | 1 | 1.1 | 0.9 | 0.9 | 0.1 | 0.9 | 0.1 | 0.99 | 0.09 |
| row 4 | 1 | 0 | 1 | 0 | 0 | 0 | 0 | 0 | 1 | 1 | 1 | 1 | 0 | 1 | 0 | 1 | 0 |

decreases toward 0
increases toward 1

*Figure 9 Softmax Gradient Calculation*

Staring with column 1, let's trace through what actually happens to the data in the network, after which, you will hopefully better understand how the three functions above actually work to move the weights toward values that calculate the results we want.

At **column 1**, we first calculate *probabilities* by running the state through a network forward pass which is shown in steps 1 & 2 above – the probabilities are the outputs of the Softmax layer.

Next, at **column 2**, the *action* is calculated by treating the probabilities as a probability distribution and selecting the action from the distribution via randomly generated number, also shown in step 3 above.

At **column 3**, the initial gradients for both action 0 and action 1 are calculated using the internal SoftmaxCrossEntropyLoss layer, which essentially subtracts the target from the Softmax output.

At **column 4**, we directly set the *Bottom.diff* to the gradients calculated in column 3.

At **column 5** in our example, let's assume that the weight value is set to 1.0.

The new weight in **column 6** is calculated with *new weight = old weight − diff*, and this actually occurs during the Solver ApplyUpdate shown in step 11 above.

To see what impact the weight update actually had, let's assume that our input data in **column 7** is the same as the probabilities that we previously had as the Softmax output, for we are trying to drive our weights to values that eventually produce the ground truths shown in rows 1 and 4.

Comparing the new result in **columns 10** and **11**, produced with *new result = data * new weight* with the old results in **column 8** and **9**, we can see that our new result values are indeed moving closer to our ground truths. In **row 2**, **column 8**, our old result of 0.1 moves to 0.09 – a little closer to the ground truth *probability* value of 0 (which would always produce an action of 1), and in **row 3**, **column 8**, our old result of 0.9 moves to 0.99 – a little closer to the ground truth *probability* value of 1 (which would always produce an action of 0).

As shown above, the SoftmaxCrossEntropyLoss layer does indeed move our weights in the direction that also moves the final result toward our ground truth values.

To see an example of the MyCaffeTrainerRL working, check out the video of Cart-Pole balancing for over a minute at [https://www.signalpop.com/examples](https://www.signalpop.com/examples).  And, if you would like to try out the MyCaffe reinforcement learning, just download the [MyCaffe Nuget](#) package for Visual Studio, or install the [MyCaffe Test Application](#) from GitHub.

The policy gradient reinforcement learning trainer source code is available on [GitHub](#).